\documentclass[sn-standardnature]{sn-jnl}

\jyear{2021}%

\theoremstyle{thmstyleone}%
\theoremstyle{thmstyletwo}%

\theoremstyle{thmstylethree}%

\usepackage{soulutf8}

\begin{document}

\title[Machine learning’s own Industrial Revolution]{Machine learning’s own Industrial Revolution}


\author*[1]{\fnm{Yuan} \sur{Luo}}\email{yuan.luo@northwestern.edu}

\author[2]{\fnm{Song} \sur{Han}}

\author[3]{\fnm{Jingjing} \sur{Liu}}

\affil*[1]{\orgdiv{Department of Preventive Medicine}, \orgname{Northwestern University}, \orgaddress{
\city{Chicago}, \postcode{60611}, \state{IL}, \country{USA}}}

\affil[2]{\orgdiv{Electrical Engineering and Computer Science}, \orgname{MIT}, \orgaddress{
\city{Cambridge}, \postcode{02139}, \state{MA}, \country{USA}}}

\affil[3]{\orgdiv{Institute for AI Industry Research}, \orgname{Tsinghua University}, \orgaddress{
\state{Beijing}, \country{China}}}


\abstract{Machine learning is expected to enable the next Industrial Revolution. However, lacking standardized and automated assembly networks, ML faces significant challenges to meet ever-growing enterprise demands and empower broad industries. In the Perspective, we argue that ML needs to first complete its own Industrial Revolution, elaborate on how to best achieve its goals, and discuss new opportunities to enable rapid translation from ML's innovation frontier to mass production and utilization.}

\maketitle

\section{Introduction}\label{sec1}

There has been extensive discussions about the potential of machine learning (ML) to catalyze the next Industrial Revolution, Industry 4.0 \citep{diez2019data,usuga2020machine}. However, compared to existing manufacturing industries already well-equipped with industry-wide standards and automated assembly lines, much of ML system development still relies on handcrafted architecture design, task-specific modeling, and manual data collection/annotation efforts. Competing frameworks including PyTorch \citep{paszke2019pytorch}, TensorFlow \citep{abadi2016tensorflow}, Caffe \citep{jia2014caffe}, Keras \citep{chollet2015keras}, Mxnet \citep{chen2015mxnet} lack interoperability support. In the research community alone, countless human efforts and computational resources have been spent on reimplementing a published method with merely a different framework. Thus, before catalyzing Industry 4.0, there is an urgent need for ML to first undergo its own Industrial Revolution. While it is a natural course for ML to have gone through a boutique and handicraft development phase, it is important to draw lessons from previous Industrial Revolutions and enlighten ML's future path. This perspective reflects on several key aspects of ML's own Industrial Revolution and how to best achieve their individual and collective goals. If successful, these actions will translate ML's innovation frontier to mass production and utilization, benefiting many enterprises across a broad range of industries. 

\section{Automated human-in-the-loop assembly network of ML systems}\label{sec2}
\begin{figure*}[t]
    \centering
    \includegraphics[width=\textwidth]{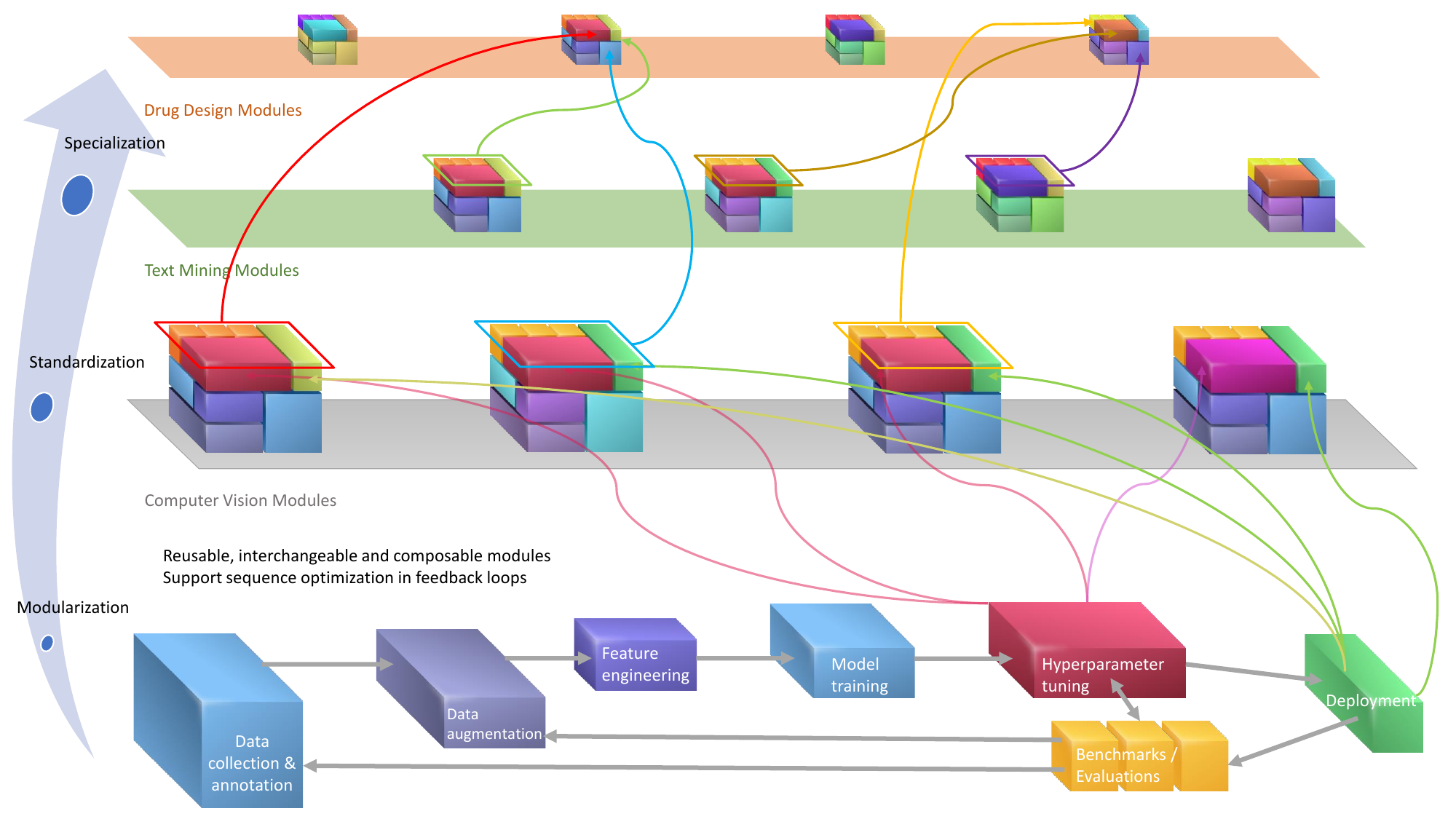}
    \caption{Machine learning's own Industrial Revolution: illustration of an automated ML assembly network. The network features distributed computing and development across developers. They construct basic and complex modules, using preferred libraries and programming languages, and collectively compose a full ML system. Moreover, they have the flexibility to iteratively customize and optimize both specific modules locally and network composition globally. Furthermore, the ML assembly network can span across multiple enterprises, pooling complex modules (e.g., computer vision and text mining) to compose modules for higher-level tasks (e.g., drug design).} 
    \label{fig:industrial-ml}
\end{figure*}

Perhaps the most well known assembly line from the Industrial Revolutions is the one developed by The Ford Motor Company that produced the Model T. Thanks to the assembly line, the Model T was produced on a mass scale, gained wild popularity and transformed the way people travel. The assembly line features modularization and standardization such that workers can perform the same task repeatedly on interchangeable parts of the car, which improves accuracy, productivity and cost-efficiency. Automated assembly lines gave humans higher-level oversight power and further freed their hands and minds for continued innovations. The adoption of the automated assembly line was a significant step forward in mass production of goods, and inspires today's ML evolution. Thanks to high-speed internet and decentralized collaboration, global ML developers can contribute instantaneously and asynchronously to the same code base, thus the organization of assembly is no longer linear but a network, hence we coin the term ML's \textit{assembly network}. Fig.~\ref{fig:industrial-ml} shows that an ML assembly network features distributed computing and consists of a diverse set of modules. Moreover, ML assembly networks enjoy the flexibility that their customization and optimization can happen both locally (node-wide) and globally (network-wide). ML assembly networks should be automated for productivity, and have humans in the loop to boost trust and incorporate domain expertise. For example, in health care, two results of a particular lab test with the same value but at different times, one in a weekday afternoon and the other at 2am on Saturday, will have different implications to a patient's status. Thus, human-in-the-loop lets domain expertise inform the ML development and application in specific industries. The notion of ML assembly network is closely related to MLOps, which aims at transitioning the algorithm to production systems, and improving and shortening the ML development life cycle. ML industrial revolution aims at reusing and sharing across many ML development life cycles and forming an assembly network, to unlock their synergies, improve network-wise industry-wide efficiency and productivity, better utilize resource and engage talent.  

\subsection{Standardization and modularization for an ML assembly network}
Standardization and modularization is key to ML assembly networks. Current ML infrastructures  already have  some off-the-shelf  basic building blocks (e.g., CNN \citep{gu2018recent}, LSTM \citep{hochreiter1997long} layers) and  sophisticated modules (e.g., VGG \citep{simonyan2014very}, ResNet \citep{he2016deep}). What is still missing, is an \textit{industrialized standardization and modularization process} so that the entire ML system can be assembled from a network of modules (Fig.~\ref{fig:industrial-ml}).  This process requires modules written in different packages (e.g., PyTorch and TensorFlow) and languages (e.g., R and Python) to be interoperable, which allows developers to compose a system using modules from different libraries and languages to leverage their complementary strengths. Moreover, shared intercommunication standards among these modules allow cross-module troubleshooting and quickly narrowing the inspection to only a few modules. 

Modularization can enable rapid iterations for both the whole system and individual modules. For example, one can start with off-the-shelf modules to quickly build a minimally-viable system, then integrate additional modules and/or refine current modules. Modularization allows effectively identifying and iterating on bottleneck modules to improve whole system performance by maximizing effects from focused efforts and resources. Module assembly can also be cast into a macro module and optimized through rapid iterations. This accelerates updating standards to match the speed of ML's evolution. With standardized modules, updating standards will often reduce to updating standards of select modules or the standards of their composition, instead of repeatedly overhauling the entire assembly network. On the other hand, we should establish industry specific standards locally to parallel generic ML standards globally. Such a hierarchical governing framework can maximally localize the standards' deliberation, hence speed up the updating cycle.  

Modularization enables distributed computing and development amongst a network of contributors. For example, for assisted driving, one organization may focus on building algorithms for scene understanding, and another may develop platforms to simulate corner cases for augmenting training data. Such \textit{vertical modularization} allows leveraging complementary strengths across organizations. On the other hand, one can also build and integrate \textit{horizontal modules}. For deployment, it is sensible to employ \textit{horizontal modularization} and let multiple local teams develop the scene-recognition modules tailored to the traffic rules and climate at different cities and countries. Vertical and horizontal modularization enables distributed computing and asynchronous development for different modules, making it feasible to recruit a large workforce with diverse backgrounds from worldwide locations.  

Modularization and standardization allow for customizations at an affordable cost. With fine-granular modularization, we only need to customize a subset of modules. Moreover, sharing and reusing common modules (Fig.~\ref{fig:industrial-ml}) further reduce amortized cost. Customizing a module can lead to many variations, abstracting them into one module class can enable efficient inventory management. Modules describable with unified mathematical functions will be instantiations of an abstract module. This allows designing ML assembly network to focus on the higher-level functions of abstract modules. The assembly network further breaks the enterprise silos in that modules from computer vision (e.g., tumor recognition from radiology images) and text mining (e.g., mining the molecular chemistry literature) can compose modules for higher-level tasks (e.g., drug design). 

\subsection{Human-in-the-loop Automation of ML assembly networks}\label{sec2.2}
Significant progress has taken place in automating ML systems, the most prominent ones including auto-sklearn \citep{feurer-neurips15a}, AutoGluon \citep{agtabular}, Google Cloud AutoML, TPOT \citep{le2020scaling}, AutoKeras \citep{JMLR:v24:20-1355}, H2O Driverless AI, Auto-PyTorch \citep{zimmer-tpami21a}, the Automatic statistician \citep{steinruecken2019automatic}, and Auto-WEKA \citep{kotthoff2019auto}. In the year 2015 AutoML systems have outperformed human experts in some of the international competitions on AutoML. For example, auto-sklearn won several of the ``Tweakathon" tracks of the first AutoML challenge organized by ChaLearn in 2015-2016 \citep{guyon2016brief}, outperforming teams of human experts who were given several months of time to obtain strong performance in a Kaggle-like competition. Since the 2013 release of Auto-WEKA, there have been 8 AutoML workshops held at the leading ML conference ICML, leading to the formation of a dedicated AutoML conference held first in 2022 (automl.cc). Meanwhile, meta-learning \citep{hospedales2021meta} that aims to enable learning to learn makes it easier to automate and generalize to new tasks, which is synergistic to AutoML. For example, it can be used to warm start hyperparameter optimization and Neural Architecture Search (NAS). Recent large language models such as ChatGPT and its precursor InstructGPT \citep{ouyang2022training} showcased progress in few-shot and/or zero-shot learning for fine-tuning the model to perform human-specified tasks.

Despite promising technical advances, considerable progresses need to take place before realizing an automated human-in-the-loop ML assembly network. For example, in NAS \citep{elsken2019neural,ren2021comprehensive}, there is not yet an NAS system that could be readily used off the shelf by citizen developers (i.e., developers with little or no programming training). While meta-learning can be used to initialize some ML modules, existing methods face significant challenges when learning on a diverse family of tasks and modules \citep{hospedales2021meta}. There also lacks shared standards and best practices across industries and scientific fields \citep{elsken2019neural}. With multiple search strategies available (e.g., reinforcement learning [RL] \citep{sutton2018reinforcement} based methods, evolutionary algorithms \citep{yu2010introduction}), and without abundant industry-specific baselines for evaluation calibration, it is difficult for citizen developers to choose from various types of NAS for domain-specific applications \citep{ren2021comprehensive}. 

Besides improving ML architecture design automation, we need to automate other ML steps (e.g., data collection) for humans to efficiently operate the assembly network. For example, ML for health care has largely been reactive to human input, but even experts may fall behind the changing disease (e.g., the mutating SARS-CoV-2 virus). Proactive ML was proposed to replace reactive ML \citep{luo2022proactive} by completing a feedback loop where ML algorithms inform humans on the data gap and guide targeted data collection. Proactive ML can also automate other steps in the ML workflow (e.g., data augmentation), and is equally applicable to non-health-care sectors. 

Besides NAS/AutoML techniques and proactive ML principles, we need to enable efficient ML system upgrades, rapid iterations, distributed development and coordinated efforts. For an effective human-in-the-loop automated ML assembly network, we envision and elaborate on the following ingredients (Fig.~\ref{fig:ml-flowchart}): 

\begin{figure*}[t]
    \centering
    \includegraphics[width=\textwidth]{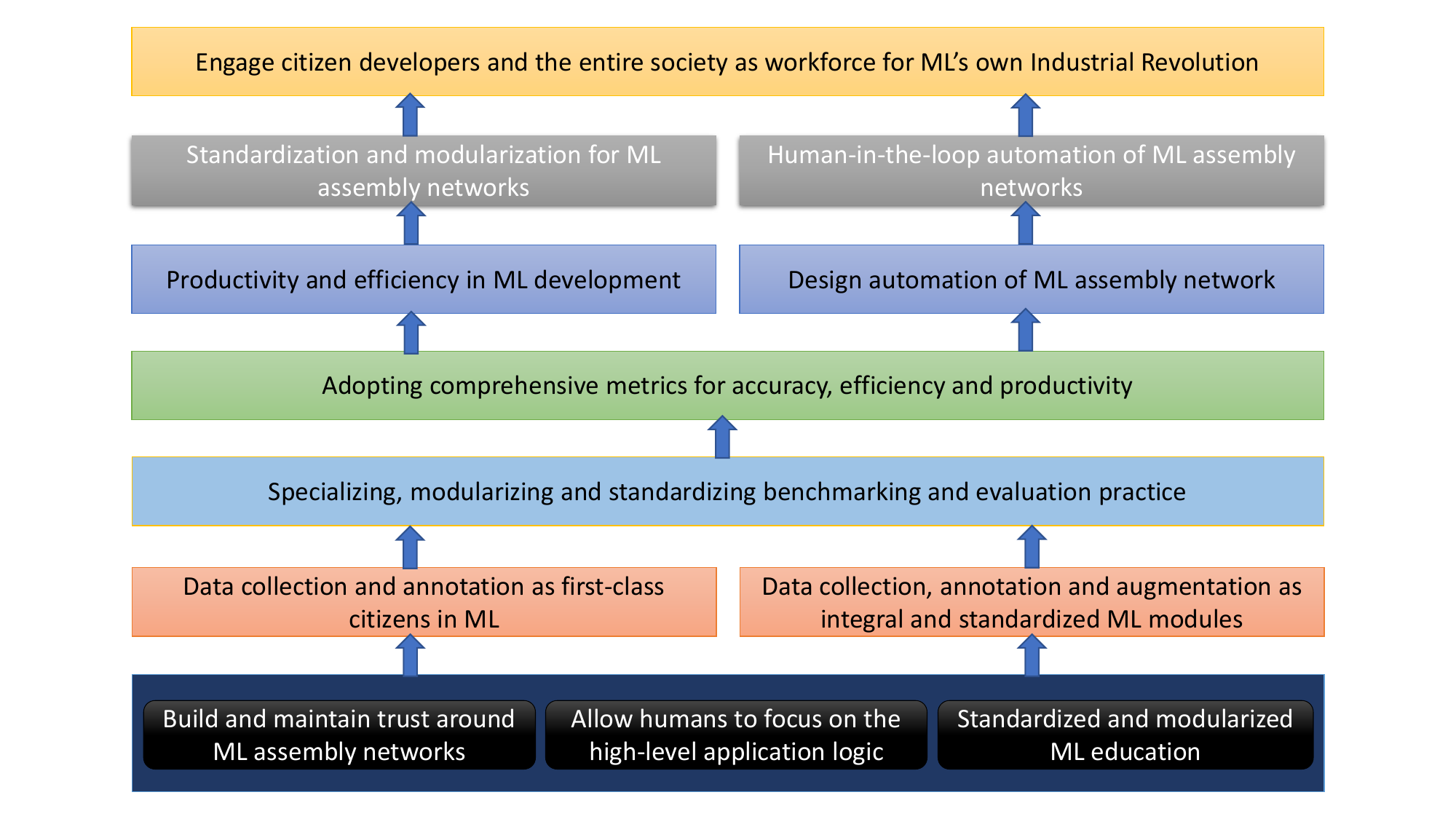}
    \caption{The different ingredients to bring together the machine learning's own Industrial Revolution.} 
    \label{fig:ml-flowchart}
\end{figure*}

\begin{enumerate}
  \item \textit{Integrated and coordinated data collection, annotation and augmentation}. (Section~\ref{sec3}) 

  
  \item \textit{Standardized evaluation tools and modularized domain-specific benchmarks}. (Section~\ref{sec4}) 
     
  \item \textit{Incorporated options for model efficiency control and cost management}. (Section~\ref{sec5}) 
  

  \item \textit{User-friendly interface for effective large-scale human-ML collaboration}. (Section~\ref{sec6}) 
\end{enumerate}


\section{Data collection, annotation and augmentation as standardized ML modules}\label{sec3}
Current ML, except for some general domain benchmarking datasets, has no universal way to create large publicly shared datasets or harden labels for data (i.e., persisting labels for features and outputs) in specific application domains. Moreover, ML models trained and tested on a single dataset that is not representative of the entire data distribution cannot reliably generalize to unseen data. This generalizability issue is especially pronounced where there are unforeseeable changes in the underlying distribution of data overtime \citep{lu2018learning}. 

\subsection{Data collection and annotation modules as first-class citizens in ML}
\begin{figure*}[t]
    \centering
    \includegraphics[width=\textwidth]{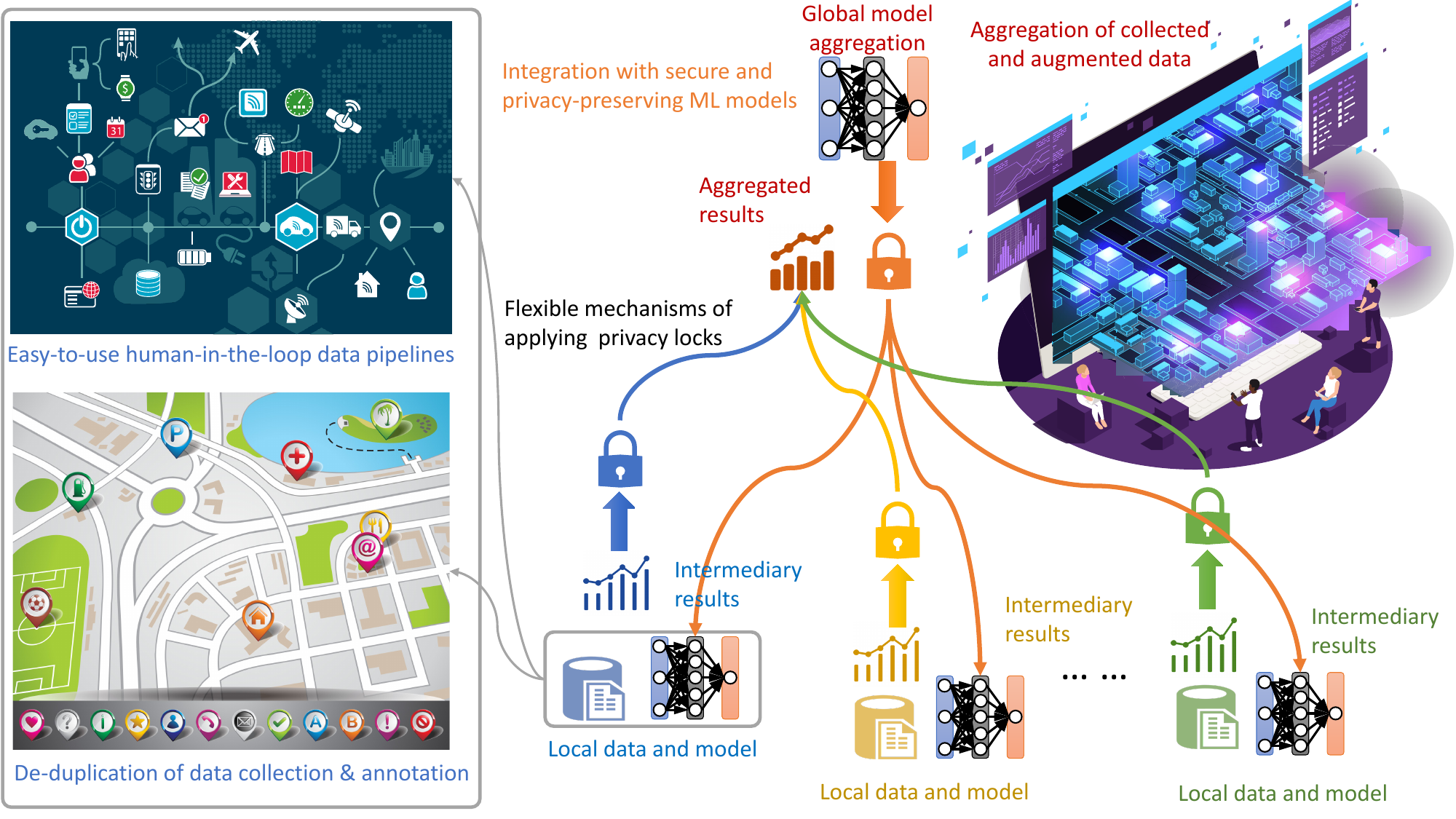}
    \caption{Data collection and annotation as first-class citizens in an ML assembly network. Shown is a federated learning setting, where example data collection and annotation modules are illustratively associated with one of the local clients and a data augmentation module with the central server. The standardization of these modules supports their flexible deployment in different parts of the ML assembly network.}
    \label{fig:data-module}
\end{figure*}

Successfully addressing the issues of data bias and data shift requires promoting \textit{data collection and annotation} as first-class citizens in ML, instead of relegating them as pre-ML steps (Fig.~\ref{fig:data-module}). If data is the oil that powers ML advancement, data collection and annotation are the critically important oil mining and refining processes. Their specifications and quality are tightly coupled with ML training, and they should be coordinated with downstream ML applications. This is analogous to different refined oil products serving corresponding needs, such as gasoline for motor vehicles and kerosene for airplanes. Collecting and annotating unstructured data effectively for meaningful use often represents a substantial challenge and is itself a complex step even using other ML modules. For example,  adding detailed companion annotations (e.g., gene regulations, diseases) to literature databases (e.g., PubMed) to support inference-style queries (e.g., searching for genes whose down-regulations are linked to a certain disease) will involve ML (NLP) modules for named entity recognition and relation extraction (e.g., genetic mutations cause diseases).  

Promoting data collection and annotation as first-class citizens in ML can complement and improve data augmentation. The research community has started to recognize the importance of data augmentation and wrangling as no less than model building, and has taken firm steps towards automated data engineering \citep{de2022automating}. Boosting data collection and annotation not only can bridge the gaps where data augmentation is less effective, but also can improve the latter for complex applications like driving-assistance. For example, enhancing sensor calibration is crucial for collecting high-quality data points, and can enable data augmentation to simulate and generate corner cases to improve driving diversity \citep{yurtsever2020survey}.  

Data collection and annotation as first-class ML citizens can better streamline quality assessment and control on data. This is particularly relevant for health care data, which is often noisy, biased and even shifting. Treating data collection and annotation as equal peers among the other ML steps can formalize quality assessment metrics including missingness, bias, and uncertainty and elucidate their impact on the overall ML system performance and generalizability. That knowledge will in turn guide quality control on data collection and annotation, and guide quality improvement using data augmentation \citep{shorten2019survey,feng2021survey,salamon2017deep}. 

\subsection{Standardizing and modularizing data collection, annotation and augmentation in ML assembly network}
Standardizing and modularizing data collection and annotation helps de-duplicate efforts and offers flexible mechanisms for fine-grained privacy control. To this end, we need common data models and universally-shared collection and annotation standards for different tenants to pool and reuse data. A common data model (CDM) provides a uniform and modular set of metadata and schemas that standardize logical infrastructure for information exchange. Shared standards unify data semantics to make ML models operate on consistent meanings across aggregated data. Modular CDMs enable fine-grained privacy control and can  maximize data utility while ensuring data security \citep{mao2018my}. For example, many patients are comfortable sharing their hospital visits for ordinary diseases (e.g., flu) but not for sensitive diseases (e.g., HIV).  Sharing visit-level instead of patient-level through modular CDMs can cover more patients. Modular CDMs also allow federated learning \citep{li2020federated} to process data at local clients for enhanced privacy (Fig.~\ref{fig:data-module}).  

Standardized data augmentation modules can enable seamless aggregation of augmented data with collected data. For example, assisted driving needs to simulate extreme weathers and corner cases that rarely occur in real-life to augment the data. Making these augmentation modules conform to universal standards of real-world data can facilitate the normalization of practices across different vendors. Modularizing data augmentation  makes it easy for humans to operate module interfaces and build human-in-the-loop process for tasks where data  need to conform to societal norms and regulations (Fig.~\ref{fig:data-module}).

Although data augmentation is a well-established field \citep{shorten2019survey,feng2021survey,salamon2017deep}, there are heterogeneous progress of standardization and modularization across application fields and data modalities. For example, text data augmentation is relatively under-explored compared to image data, and lacks consensus over the best practice \citep{feng2021survey}. Even for computer vision, despite standardization and modularization of common augmentation functions in TorchVision and progress in building automatic augmentation policies and establishing data augmentation inside AutoML, more advanced techniques such as conditional GANs \citep{mirza2014conditional} and neural style transfer \citep{gatys2016image} may still exist as boutique implementations awaiting to be reworked into standardized and modular  libraries.


Data collection/annotation/augmentation modules should also support the \textit{customized specialization of generic algorithms} in different application fields. Following the computer vision example, there are few if any standard ML library implementations tailored for medical imaging data augmentation. For example, chest X-rays from patients in the same age group or from the same viewing angles can inform the classification of disease  conditions \citep{mao2022imagegcn}. Such relations among X-rays should be leveraged by specialized data augmentation and packed into a new module for medical imaging, to answer for widespread demands from scientists and clinicians. They can in turn contribute to devising specific quality control criteria to guide the continuous development and refinement of the shared augmentation modules. Thus, specializing data collection/annotation/augmentation modules into individual industry is important in order to make more tangible impact. Due to the significant undertakings, this is hard to achieve if solely relying on open source initiatives. Having industry professionals collaborate with the open-source community can help create a mixture of open source tools and managed enterprise solutions and effectively ground them for industry tasks. 


\section{Specializing, modularizing and standardizing benchmarking and evaluation practice}\label{sec4}
With the advancement of AutoML, considerable progress has been made in building end-to-end benchmarks including the OpenML AutoML benchmarks \citep{amlb2019}, the AutoML challenges' benchmarks \citep{guyon2019analysis} and AutoDL challenges' benchmarks \citep{liu_winning_2020}, which began to cover application areas such as biology and medicine, energy and sustainability management. Progresses are also seen in modular benchmarks on subsystems like neural architecture search, e.g., tabular and surrogate NAS benchmarks \citep{mehta2022bench}, and hyperparameter optimization (e.g., HPObench \citep{eggensperger2021hpobench}, HPO-B \citep{arango2021hpo}, or JAHS-Bench-201 \citep{bansal2022jahs}). From these benchmarks and challenges, best practices on model comparison and reporting were summarized, such as in \citep{JMLR:v21:20-056}. 

Despite the progress, there are still unmet needs in specializing ML benchmarking and evaluation process to successfully deploy applications to domain-specific tasks. For example, pharmaceutical industry aims to generate as many quality drug candidates as possible, as quickly as possible, with the highest probability of successful transition to clinical development. In contrast to many conventional ML problems that aim to improve average results to outperform state-of-the-art methods on existing benchmarking datasets, ML for drug discovery is a best-case problem with multipronged objectives. When existing domain-specific benchmarks are available, they are often limited in quantity, size and diversity. For example, in health care, despite significant progresses such as MIMIC \citep{johnson2016mimic}, TCGA \citep{cancer2013cancer}, UK Biobank \citep{sudlow2015uk}, many diseases still have no publicly available benchmark datasets for ML experiments. Even for diseases where public benchmarks do exist, they are typically scattered without consistent data collection and annotation, and smaller in scale than generic-domain benchmarks. Addressing these unmet needs, ML communities such as NeurIPS launched tracks on datasets and benchmarks that started in 2021 \citep{vanschoren2021announcing}, which are gaining momentum. However,  it will take some time to build systematic infrastructure across all major industries. 


 
Specialized benchmarks often need to evaluate multiple dimensions of objectives, and should leverage modularization and standardization for better scaling. For example, applications may need to evaluate growing dimensions of objectives such as bias and equity and an associated set of metrics. Modularized evaluation that is extensible better suits the evolving application field, enables reuse and saves efforts.  
To make the modular evaluation reproducible, we need to standardize a consensus set of evaluation dimensions for unit benchmarks in the specific industry. As an analogy, airplane quality inspection employs specialized (e.g., tailored for different types of airplane), modularized (e.g., unit testing for different parts) and standardized (e.g., meeting aviation standards) evaluation. Similarly, modularization and standardization of ML benchmarking can benefit from dividing lengthy processes into standardized steps of unit testing and integration testing, for better efficiency and precision. 

As ML tasks grow increasingly complex, the review process of ML publications, which are ML's innovation frontier and have increasingly emphasized reproducibility and objective benchmarking, should put more emphasis on modularization and standardization. For example, we need to  assign dedicated reviewers on different benchmarking aspects, similar to statistics reviewers in leading medical journals. The inaugural AutoML conference 2022 introduced reproducibility reviewers, which set a pioneering example for other venues and the field. Evaluation configurations need to be standardized for fair comparison. A simple example is p-value, an indicator for statistical significance, which can be quite different and even on opposite sides of the significant threshold when comparing the same two ML systems with datasets of similar content but different sizes. It is advisable to precisely specify the context for consistent calculations and interpretations of p-values for standardized review. Impartial third-party inspectors who are specialized in specific modules of an application field can further help standardization and modularization. These evaluation authorities can apply \textit{product labels} for quality assessment and control of data collection/annotation/augmentation and other modules of ML systems, clearly stating their features and quality ratings analogous to regular vs. premium gas in the oil industry. These labels can guide ML modules to be successfully integrated in the assembly network and function properly. They can also contain hazard warnings such as bias and failure cases, to guide ML systems' continuous evaluation after deployment. 

\section{Adopting comprehensive metrics for efficiency and productivity}\label{sec5}

\subsection{Computational efficiency in ML development}
With growing hardware-aware ML architectures that take into account latency and memory constraints, computational efficiency of ML systems should also be prioritized. Training gigantic ML models consumes astronomical  storage, memory bandwidth and computational resources. For example, the GPT-3 model \citep{brown2020language} contains 175 billion parameters and takes 355 GPU years to train for \$4.6M in cloud computing cost. Such prohibitive and non-sustainable expenses make it difficult to democratize ML to economically disadvantaged populations. Extensive model training also causes severe environmental problems such as high carbon emissions \citep{strubell2019energy}.  

Edge ML is one promising avenue for efficient ML, which targets deploying ML on edge devices including mobile phones and smart vehicles, and brings the benefits of lower cost, lower latency, and better privacy. However, performing resource-intensive inference on edge devices is challenging \citep{cai2022enable}. For instance, a state-of-the-art machine translation model requires more than 10 billion multiply-and-accumulates (MACs) to process a sentence of only 30 words. Such a high computational cost is far beyond the capabilities of most mobile devices, since their hardware resources are tightly constrained by the form factor (the physical size and shape of a device), battery, and heat dissipation. These computation workloads, however, cannot be delegated to the cloud server as they are sensitive to latency (e.g., autonomous driving) and/or privacy (e.g., healthcare). Therefore, efficient deep learning is important for lifting the roadblock for Edge ML applications.  

\begin{figure*}[t]
    \centering
    \includegraphics[width=\textwidth]{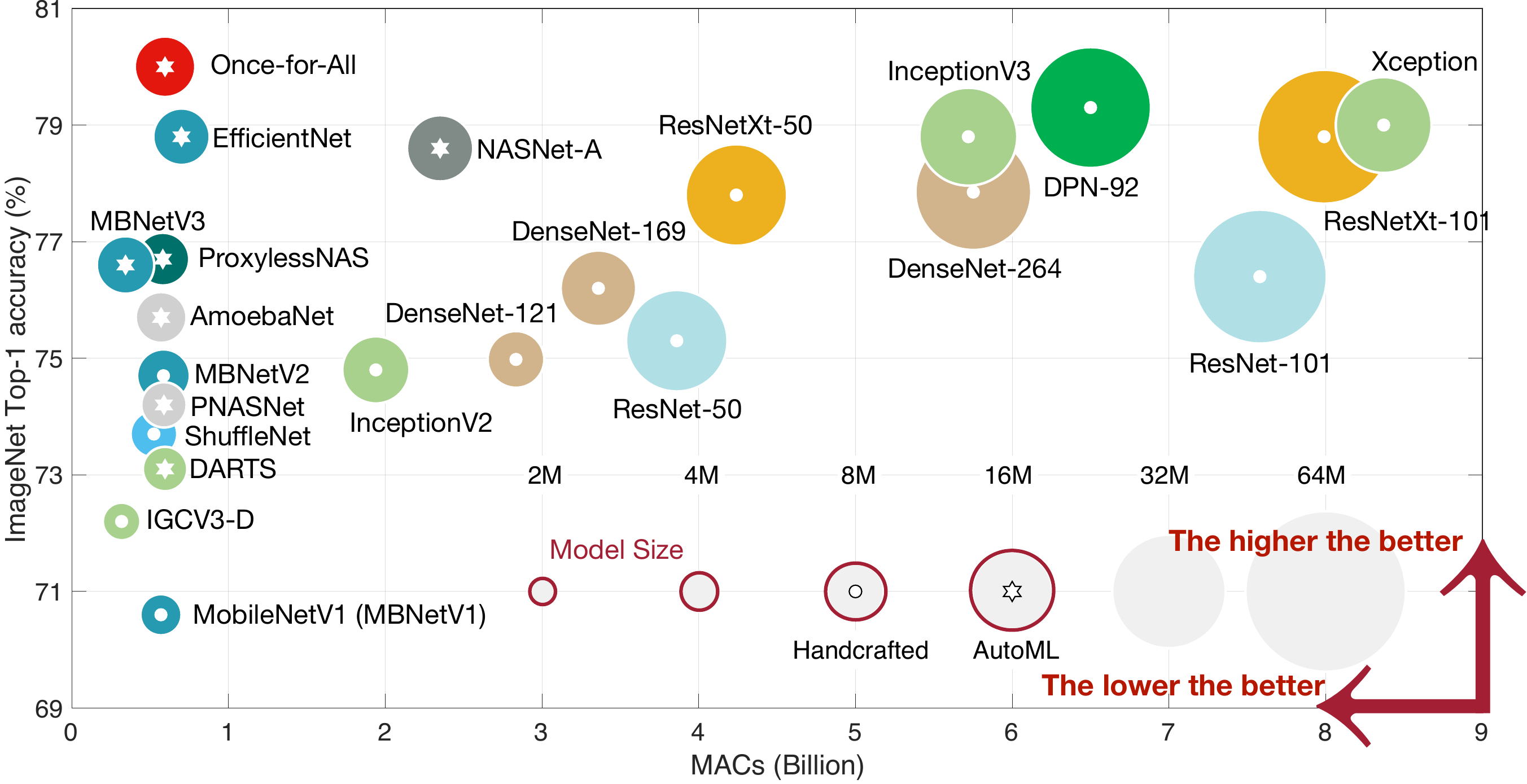}
    \caption{Multiply-and-accumulates (MACs) vs accuracy for popular neural networks. Observation 1: With the same accuracy, there is a big efficiency gap among different models. Observation 2: AutoML designed models outperform human design and achieve a better accuracy-computational cost balance. Image adapted and updated from \citep{deng2020model}.}
    \label{fig:automl}
\end{figure*}

It is critical to innovate efficient ML techniques, such as pruning \citep{han2015learning,frankle2018lottery}, model compression \citep{han2016deep}, knowledge distillation \citep{hinton2015distilling}, and specialized hardware design \citep{han2016eie}, to create a more sustainable ML landscape. Suggested metrics for efficiency, greenness and sustainability of ML models are listed in Table~\ref{tab:metrics}. An ideal ML model would have small model size and peak memory footprint, minimal inference latency and MACs, high training and inferencing throughput, and require low energy consumption. These goals are not all compatible among themselves and with other performance metrics (e.g., accuracy). However, as Fig.~\ref{fig:automl} illustrates, it is possible to find a sweet spot to balance the multidimensional objectives for computationally efficient and sustainable ML development. 

\begin{table}
\centering
\caption{Metrics for creating a greener and more sustainable ML landscape}
\begin{tabular}{l|l}  
\toprule
Metric & Definition \\
\hline
Model Size & The number of parameters times the bitwidth for each parameter. \\
Inference Latency & The time to perform inference given certain batch size. \\
Inference Throughput & The number of inferences per second. \\
Inference MACs & The number of multiplication and accumulations (MACs) per inference (usually single batch size). \\
Peak Inference Memory & The maximum memory consumption during inference. \\
Training Throughput & The number of training data per second. \\
Training Time & The time it takes from the training starts till convergence. \\
Training Memory & The memory consumption during training. \\
Energy Consumption & The energy consumption for training or inference. \\
\bottomrule
\end{tabular}
\label{tab:metrics}
\end{table}

\subsection{ML development productivity and design automation of ML assembly network}
ML assembly network requires significant design efforts for individual nodes and their interactions so that the resulted models perform well. To strike the right balance between accuracy, efficiency and hardware constraints, we need to inspect the ML assembly network with the prismatic lens of multiple requirement dimensions to maximize the overall performance, which further adds to the design burden and imposes challenges on ML development productivity.  

Automated design methodology are needed to assist users in optimizing their ML architectures along  dimensions such as accuracy, efficiency and hardware constraints. Conventionally, designers use their expertise to design efficient neural network architectures such as MobileNets \citep{howard2017mobilenets}, ShuffleNets \citep{zhang2018shufflenet}, and CondenseNets \citep{huang2018condensenet}. These techniques heavily depend on experts’ hand-crafted and heuristic-driven designs. However, only a limited number of people in academic and industrial organizations possess the ability to create advanced full-stack machine learning systems that span feature learning, optimization, parameter tuning, and hardware deployment. Hardware-aware NAS \citep{benmeziane2021hardware,li2021hw} attends to both accuracy metrics and hardware-aware constraints, and has the potential to further optimize their tradeoffs, as shown in Fig.~\ref{fig:automl}.  Challenges arise for recent large language models (e.g., ChatGPT) and diffusion models (e.g., DALL-E), because the training cost is too high, and conventional model compression techniques (e.g., pruning, NAS) heavily requires retraining. One potential remedy is to develop training-free or training-lite model compression techniques for these emerging models. 

To further boost productivity and scalability, we can leverage meta-learning \citep{hospedales2021meta}, pre-training and foundation models \citep{bommasani2021opportunities} to accelerate streamlining  module specialization  while reducing the training cost for diverse downstream tasks. These techniques can amortize training costs and make models adaptive and reusable for different tasks. When adapting foundation models in the ML assembly network, one challenge is how to efficiently select the right module scope and the right model capacity for different industries and tasks. Addressing this industry-dependent challenge will necessarily involve not only ML model research and benchmarking, but also the associations and governing bodies of said industries to carve out module scope (e.g., an entire dialogue system or named entity recognition module) and enact bespoke standards for that industry (e.g., on common data model and annotation). 

Maximizing the automation of individual nodes and their interactions in ML assembly networks requires expansions of AutoML that are both deeper (e.g., automating building block design) and wider (e.g., adding more steps and performing network-level optimization). Such a task needs to keep humans in the loop and will be inherently complex and necessitate collaborations with other fields. For example, human computer interaction (HCI) principles can help design a control panel and dashboard with proper complexity to enable effective human control. Industrial organization and operations research should be employed to optimize the workflow (e.g., distributing and scheduling a mixture of jobs with different computation or human resource overhead) across the ML assembly network. Such a collaborative automation and optimization approach can serve a variety of industrial applications such as supply chain, finance, education and transportation. This makes ML experts focus on advancing new fields in ML, helps less-skilled citizen developers quickly warm up to the task, and allows businesses with limited ML expertise to build their own high-quality bespoke models more effectively. 

\section{Rethinking human's role in ML assembly networks}\label{sec6}
Recent large language models such as ChatGPT makes it feasible for ML to take over the role of coding, allowing humans to focus on formulating intent and output logics to guide the code generation for ML modules and systems. These advances bring the vision of ML Industrial Revolution closer to reality, and prompt rethinking human's role in ML assembly networks.

First, we need human-in-the-loop automation to build and maintain trust around ML assembly networks. Standardized modules with straightforward functions make it easier for humans to oversee quality checking and A/B testing to demonstrate their head-to-head advantages, which boosts confidence. For example, data augmentation/wrangling modules with specialized industry experts in the loop can ensure they meet enhanced industry-specific standards. Continuous, standardized and modularized evaluation help keep the data and model up to date and more reliably serve our needs. These all bolster trust. Moreover, modularization allows separating the modules that need interpretation with those that do not. We can then focus on better improving the interpretability of needed modules and keep them under more scrutiny to further boost trust. In addition, when automated ML assembly networks may encounter unseen scenarios and take erroneous actions by themselves, human-in-the-loop interventions can be an extra safeguard to ensure human accountability and add trust to the process.

Second, we need to build user-friendly interfaces to support effortless human-module interaction, and seal the module implementation under the hood, to allow users to focus on the high-level application logic. The assembly network breaks down a complex task into subtasks and automate them to relieve professionals from having to master them manually. This allows humans to attend to increasingly higher-level tasks, and stay in the loop for allocating resources, designing processes and laying out pipelines. For ML assembly networks, we need to embrace the added multidimensional complexity  beyond the linear progression, and ponder over the questions including: Should humans function as a commander or a \textit{collaborator/co-pilot} to the ML system? Given successes such as ChatGPT, should humans reverse-adapt to ML (e.g., new prompt engineering) instead of only ML adapting to humans? Can we better explore standardization, specialization and modularization to sustain \textit{human-ML co-adaptation} in the quest for human-in-the-loop ML assembly network?  



Third, ML education needs to be standardized and modularized to make increasingly specialized expertise interoperable. Standardization and modularization enable citizen developers to focus on particular modules, and automation further reduces knowledge requirement to basic conceptual level. Existing ML education lacks dedicated modular and in-depth courses on data generation and augmentation, ML ethics, evaluation methods, efficient deployment, and bias issues. Providing standardized modular courses in ML education will give people the flexibility to choose only the needed modules and reduce time and resource commitment. This will make human-ML co-adaptation more feasible, through divide-and-conquer to lower the bar. These modules should all fit into a shared underlying framework to promote interoperability. Coupling with credential certifications can ensure the qualification of graduated trainees. This is analogous to training workers to fill different roles in a factory, and can more efficiently disseminate and democratize ML literacy, engage the entire society as workforces, and achieve human-ML co-adaptation at scale.  


\section{Conclusion}\label{sec7}
In the modern world, enterprises across multiple industries seek to leverage ML to draw intelligence from their ever-growing datasets and inform daily operations and long-term strategies. However, without the support of universally accepted standards and human-in-the-loop automated assembly network, much of the ML practice still relies on handcrafted designs and task-specific manual data collection and annotation processes, making it difficult to scale up and meet diverse enterprise demands. Before it can truly enable Industry 4.0, ML development needs to complete its own Industrial Revolution to establish a human-in-the-loop automated ML assembly network. These substantial undertakings require rethinking human's role as ML collaborator/co-pilot, and reshaping education practice to cultivate a societal workforce and drive human-ML co-adaptation. By breaking ML development down to standardized and modularized steps across different industries, and simplifying them into routine actions that can be internalized into our muscle memory, a human-in-the-loop automated ML assembly network can in turn power Industry 4.0 on the horizon. 

\backmatter







\noindent

\bigskip




\bibliographystyle{unsrtnat}
\bibliography{sn-bibliography} 


\end{document}